\documentclass[10pt,twocolumn,letterpaper]{article}

\usepackage[pagenumbers]{cvpr} 

\usepackage{graphicx}
\usepackage{amsmath}
\usepackage{amssymb}
\usepackage{booktabs}
\usepackage{color}
\usepackage{color}

\usepackage{algorithm}
\usepackage{algorithmic}
\usepackage{multirow}
\usepackage[pagebackref,breaklinks,colorlinks]{hyperref}

\usepackage[capitalize]{cleveref}
\crefname{section}{Sec.}{Secs.}
\Crefname{section}{Section}{Sections}
\Crefname{table}{Table}{Tables}
\crefname{table}{Tab.}{Tabs.}

\def\cvprPaperID{5962} 
\def\confName{CVPR}
\def\confYear{2022}

\begin{document}


\title{
End-to-End Semi-Supervised Learning for Video Action Detection
}  

\author{Akash Kumar  \qquad   Yogesh Singh Rawat\\
Center for Research in Computer Vision\\
University of Central Florida\\
{\tt\small akash\_k@knights.ucf.edu  yogesh@crcv.ucf.edu}
}
\maketitle

\begin{abstract}

In this work, we focus on semi-supervised learning for video action detection which utilizes both labeled as well as unlabeled data. We propose a simple \textbf{end-to-end consistency based} approach which effectively utilizes the unlabeled data. Video action detection requires both, action class prediction as well as a spatio-temporal localization of actions. Therefore, we investigate two types of constraints, \textbf{classification consistency}, and \textbf{spatio-temporal consistency}. The presence of predominant background and static regions in a video makes it challenging to utilize spatio-temporal consistency for action detection. To address this, we propose two novel regularization constraints for spatio-temporal consistency; 1) \textbf{temporal coherency}, and 2) \textbf{gradient smoothness}. Both these aspects exploit the \textbf{temporal continuity} of action in videos and are found to be effective for utilizing unlabeled videos for action detection. 
We demonstrate the effectiveness of the proposed approach on two different action detection benchmark datasets, UCF101-24 and JHMDB-21. In addition, we also show the effectiveness of the proposed approach for video object segmentation on the Youtube-VOS which demonstrates its \textbf{generalization capability}.
The proposed approach achieves competitive performance by using merely \textbf{20\%} of annotations on UCF101-24 when compared with recent fully supervised methods. On UCF101-24, it improves the score by \textbf{+8.9\%} and \textbf{+11\%} at 0.5 f-mAP and v-mAP respectively, compared to supervised approach. The code and models will be made publicly available at: \url{https://github.com/AKASH2907/End-to-End-Semi-Supervised-Learning-for-Video-Action-Detection}.

\end{abstract}

\section{Introduction}
\label{sec:intro}

We have seen a great progress in video action classification \cite{i3d, action_recog1, action_recog2, action_recog4, action_recog3, vyas2020multi, demir2021tinyvirat, tirupattur2021modeling}, where the availability of large-scale datasets is one of the enabling factor \cite{jhmdb, ucf101, k400}. Video action detection on the other hand is much more challenging where spatio-temporal localization is performed on the video. In addition, obtaining large-scale datasets for this problem is even more challenging as annotating each frame is a huge time and cost intensive task.

\begin{figure}[t]
     \centering
         \centering
         \includegraphics[width=0.235\textwidth]{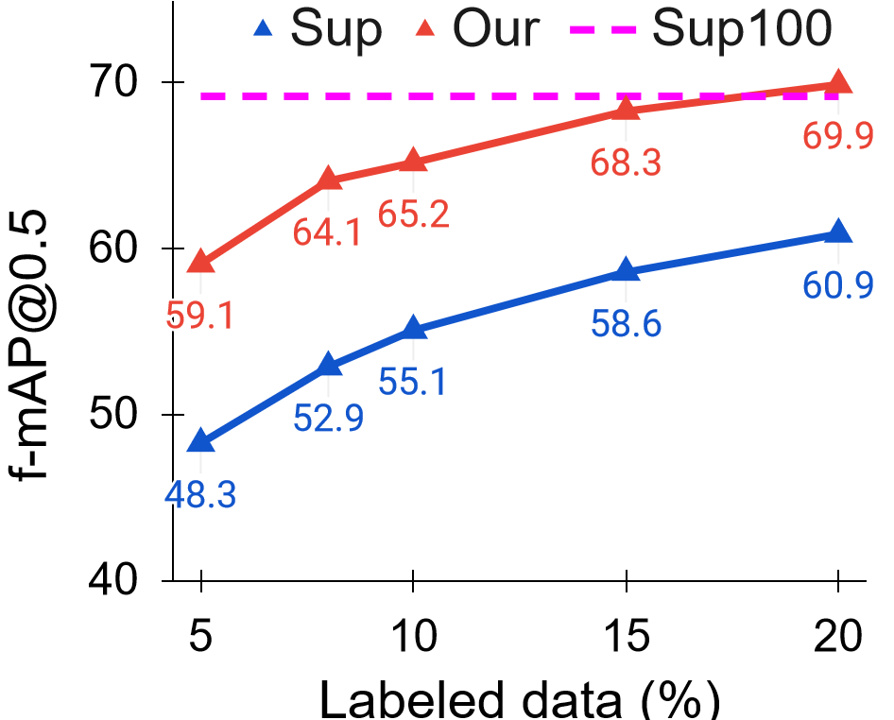}
         \includegraphics[width=0.235\textwidth]{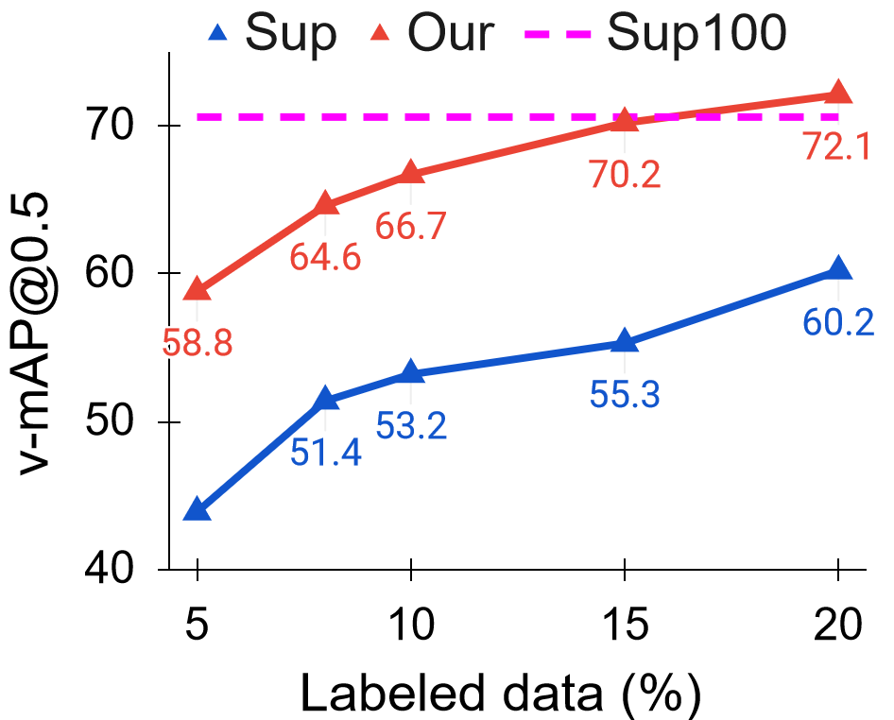}
     \caption{A comparison of proposed semi-supervised method with supervised baseline showing absolute gain in f-mAP and v-mAP for varying number of labeled samples on UCF-101-24 dataset. The proposed method outperforms supervised baseline and using merely 20\% of labeled samples, matches the performance of fully supervised method trained on 100\% labels. Sup is supervised and Sup100 is supervised with with 100\% labels. 
     }
     \label{fig:varypercentage}
\end{figure}

In this work, we focus on semi-supervised learning for video action detection which makes use of a small set of annotated samples along with several unlabeled samples.
For annotated set, we have video-level class labels as well as frame-level localizations. To the best of our knowledge, this is the \textit{first work} which focuses on semi-supervised learning for video action detection.

Semi-supervised learning has been successfully studied for image classification \cite{mixmatch, uda, fixmatch} with some recent works in object detection \cite{co_ssd, semiobj1, semiobj2, semiobj3, semiobj4}. Pseudo-labeling \cite{pseudocls, rizve2020defense}  and consistency regularization \cite{fixmatch, uda, semiobj1} are two main approaches used for semi-supervised learning. Where pseudo-labeling rely on several iterations, consistency regularization relies on single-step training. Since, training a video action detection model is already \textit{computationally expensive} due to high-dimensional input, therefore we propose a \textit{consistency-based} approach for an efficient solution.  

Video action detection requires a sample level class prediction as well as a spatio-temporal localization on each frame. Therefore, we investigate two different consistency constraints to utilize unlabeled samples; \textit{classification consistency} and \textit{spatio-temporal localization consistency}. Consistency regularization for classification has been found very effective \cite{mixmatch, fixmatch}, however, it relies on a rich set of augmentations. Extending these augmentations to the video domain for spatio-temporal consistency is not always feasible. 

We propose a simple formulation for spatio-temporal consistency where it is computed for each pixel in the video. Extending traditional consistency objective to spatio-temporal domain could capture pixel level variations, but it fails to capture any \textit{temporal constraints} as the consistency is computed independently for each pixel.
To address this issue, we explore \textit{temporal continuity} of actions in videos. We argue that motion has some temporal continuity and we attempt to utilize this to regularize the spatio-temporal consistency. We investigate two different ways to capture motion continuity, \textit{temporal coherence} and \textit{gradient smoothness}. Temporal coherence aims at refining the uncertain boundary regions that distinguish foreground and background, and, gradient smoothness enforces temporally consistent localization.

The proposed method is trained end-to-end utilizing both labeled and unlabeled samples without the need for any iterations which makes it efficient. 
We demonstrate its effectiveness with an extensive set of experiments on two different datasets, UCF101-24 and JHMDB-21. 
We show that with \textit{limited labels} it can achieve competitive performance when compared with \textit{fully-supervised} methods outperforming all the \textit{weakly-supervised} approaches. In addition, we also demonstrate the \textit{generalization capability} of the proposed method on Youtube-VOS for video object segmentation. We make the following contributions in this work,
\begin{itemize}
\setlength\itemsep{-.3em}
    \item We propose a simple \textit{end-to-end approach} for semi-supervised video action detection. To the best of our knowledge, this is the \textit{first} work focusing on this problem.
    \item We investigate two different consistency regularization approaches for video action detection; \textit{classification consistency} and \textit{spatio-temporal consistency}.
    \item We propose two novel regularization constraints for spatio-temporal consistency, \textit{temporal coherency} and \textit{gradient smoothness}, which focus on the \textit{temporal continuity} of actions in videos.
\end{itemize}

\section{Related Work}

\paragraph{Video Action Detection}

Video action detection has made significant progress in recent years \cite{Saha_2016, Peng_2016, Singh_2017_ICCV, step_cvpr, Zhao_2019_CVPR, li2020actions, Zhao2021TubeRTF}, which is mainly attributed to convolutional neural networks.  Earlier attempts start from 2D proposals and then it moves towards 3D proposals, the authors in \cite{tube_cnn} extend the 2D proposals to 3D cuboids for locating actions in videos. Similarly, \cite{kalogeiton_2017} utilizes a sequence of frames, and, outputs an anchor cuboid for action localization. In \cite{step_cvpr}, the authors propose to update rough proposals progressively during the training that proves to be effective. To take advantage of a longer temporal sequence, the authors in \cite{Song_2019_CVPR} utilize a recurrent approach with the help of a Conv LSTM. Some approaches also rely on optical flow \cite{Zhao_2019_CVPR, ava}, however, it incurs additional computational cost. 

Most of these existing methods utilize a proposal-based approach \cite{tube_cnn, step_cvpr, Zhao_2019_CVPR, ava} which requires a two-step process and makes these methods complex. In this work, we utilize a simple architecture as an action detection network, which is an end-to-end approach based on capsule routing \cite{duarte2018videocapsulenet}. Although the authors in \cite{duarte2018videocapsulenet} propose a simple architecture, the requirement of 3D routing makes it computationally expensive. Therefore, we use a modified model as our baseline action detection network and utilize a 2D routing \cite{sabour2017dynamic} instead to make it computationally efficient.

\paragraph{Weakly-supervised action detection}
Video action detection requires annotations on every frame of a video for localization. To alleviate this high annotation cost, recently some weakly-supervised approaches have been proposed \cite{cheron_2018, Escorcia2020GuessWA, Arnab2020UncertaintyAwareWS}. In \cite{cheron_2018}, the authors explore the impact of different levels of supervision for action detection. The authors in \cite{Escorcia2020GuessWA} utilize siamese similarity over frames and localize actions with the help of actor proposals generated by object detectors. Similarly, the authors in \cite{Arnab2020UncertaintyAwareWS} detect actions using an off-the-shelf human detector trained on image datasets \cite{coco} with the help of multiple instance learning. Although weakly-supervised approaches reduce the annotation cost on every frame, the performance of these methods is still far from fully supervised approaches. Moreover, it requires class labels for all the samples and also relies on additional bounding boxes localization from state-of-the-art detectors such as Detectron \cite{Detectron2018} and Faster-RCNN \cite{ren2015faster}.

\paragraph{Semi-Supervised Learning}

Semi-supervised learning utilize a finite number of labeled samples along with a large number of unlabeled samples. Pseudo-labeling \cite{pseudo_obj1, pseudocls, rizve2020defense} is an iterative approach that makes it computationally expensive and not well suited for video action detection. Consistency regularization makes use of perturbations on input data and attempts to minimize the difference between predictions from augmented versions of the same sample \cite{fixmatch, mixmatch, remixmatch, survey_semi, co_ssd}. Since it doesn’t require multiple iterations, it is efficient in comparison with pseudo-labeling.

\begin{figure*}
    \centering
    \includegraphics[width=0.9\linewidth]{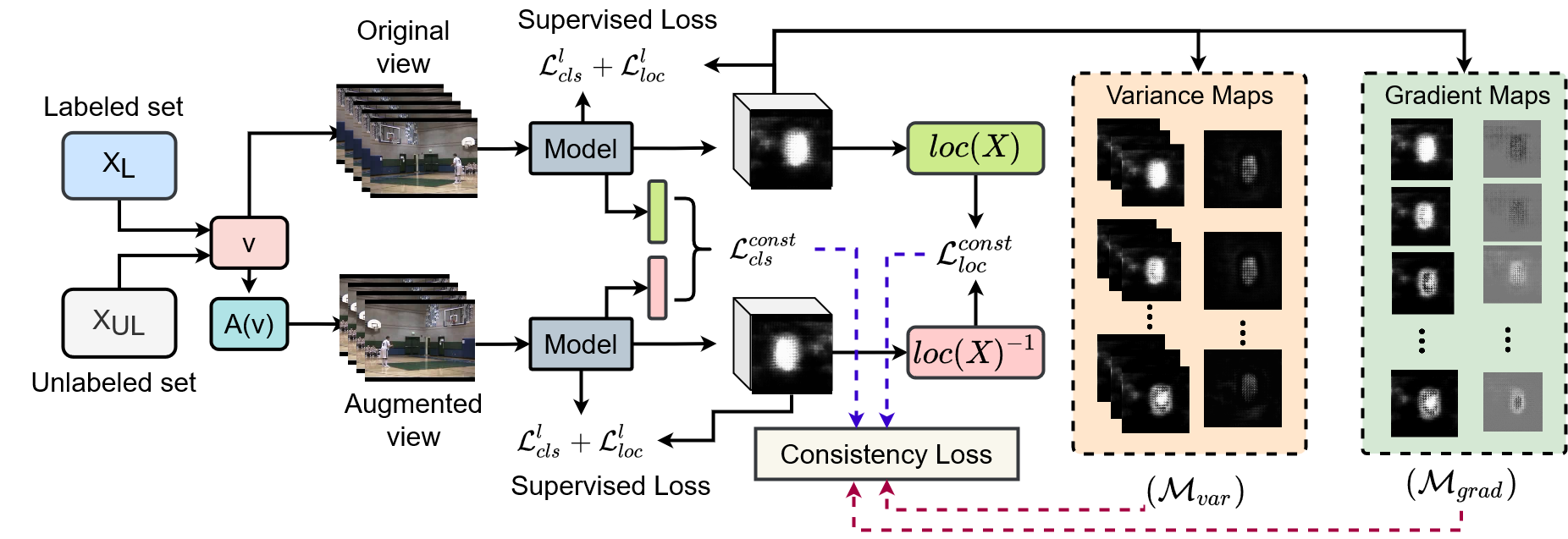}
    \caption{Overview of our proposed approach. Original and augmented view of the input video is passed through the network. The activations at the penultimate layer of classifier head are considered for classification consistency and the spatio-temporal localization is considered for localization consistency. The \textit{attention} masks $\mathcal{M}_{var}$ and $\mathcal{M}_{grad}$ are computed for temporal coherence and gradient smoothness using the spatio-temporal localization. In addition, traditional supervised classification and localization loss is computed for the labeled samples. 
    }
    \label{fig:architexture}
\end{figure*}

Recently, these approaches have been explored for semi-supervised action recognition \cite{semi_action1, semi_action2, rizve2020defense} and image object detection \cite{co_ssd, semiobj1, semiobj2, semiobj3, semiobj4}. \cite{semi_action2} keeps an action bank trained on UCF-50 activities whose features are further utilized for the semi-supervised training. \cite{semi_action1} introduces a  group contrastive loss to boost the classification score. However, the later one requires multiple stage of training, and, the previous one uses weights pretrained from same classes.  For object detection in images, most of the work follows setup for the teacher-student network. Inspired by the simplicity of $\pi$-model consistency-based approaches and its success in classification and object detection, we propose a consistency-based approach for video action detection. Also, there is \textit{no existing} work on semi-supervised video action detection to the best of our knowledge.

\section{Approach}

Given a video $v=(v_{1},v_{2}..., v_{n})$ with $n$ frames, we want to perform spatio-temporal localization which provides a class label $p$ for the whole video and localization map $l$ on each frame $v_{i}$. Localization map $l$ can be pixel-wise prediction \cite{jhmdb}  or a bounding-box \cite{ucf101}. In semi-supervised learning, the dataset is consists of a labeled (${D}_{L}$) and an unlabeled (${D}_{UL}$) set. Let's denote the whole training set with $X$, labeled subset as $X_{L}: \{ v_{l}^{0},  v_{l}^{1}, ..., v_{l}^{N_{l}}\}$ and unlabeled subset as $X_{U}: \{ v_{u}^{0},  v_{u}^{1}, ..., v_{u}^{N_{u}}\}$. We want to utilize both these sets to train an action detection model $M$.

Each training sample $v$ is augmented to get a second view $v^{'} (A(v))$.  The action detection model $M$ is used to predict a class label and spatio-temporal localization $cls$, $loc$ = $M(v)$  for each sample $v$. A traditional supervised loss is computed for classification $(\mathcal{L}_{cls}^{l})$ and localization $(\mathcal{L}_{loc}^{l})$ for a labeled sample. We utilize consistency regularization for both labeled and unlabeled samples. We calculate the difference between a sample $(v_{u})$ and its augmented view $(v^{'}_{u})$ for consistency. We investigate two different consistency loss for action detection, classification $(\mathcal{L}_{cls}^{const})$ and spatio-temporal $(\mathcal{L}_{loc}^{const})$. An overview of the proposed approach is shown in figure \cref{fig:architexture}. Next, we go through in detail about the action detection model $M$ and these two consistency regularization loss terms.

\subsection{Action detection model}
We propose a simple action detection model $(M)$ based on VideoCapsuleNet \cite{duarte2018videocapsulenet}. VideoCapsuleNet is a 3D convolution based encoder-decoder architecture. It utilizes spatio-temporal features for detecting and localizing actions in a video. Although it is a simple architecture, the use of 3D capsule routing increases the computation overhead significantly. We propose to use 2D routing \cite{2drouting}  instead of 3D routing after pooling the temporal dimension of features and found it to be more efficient without much performance drop. We utilize this adapted model in our experiments. This model $M$ provides a classification prediction $p$, and, a spatio-temporal localization $l$ for an input video.

\subsection{Classification consistency}

We want the classification prediction for a sample and its augmented view to be similar. We looked into the output of the latent features of the original view ${feat(X)}$ and the augmented view $feat(X^{'})$ from the network. The intuition is that the variation in the distribution should be minimal. To enforce this, we employed Jenson-Shannon divergence (JSD) to compute the difference between them.
Using JSD, the classification consistency loss $(\mathcal{L}_{cls}^{const})$ is defined as:
\begin{equation}
    \label{eqn:clsconst}
    \mathcal{L}_{cls}^{const} = \mathcal{L}_{JSD} =  JSD(feat(X), feat(X^{'})).
\end{equation}

\subsection{Spatio-temporal consistency}
In this consistency constraint, the network learns to detect spatio-temporal localization for multiple views of a video. Using a sample $(v)$, the action detection network $(M)$ outputs a localization map $(l(v))$, which is a pixel-wise prediction, where each pixel has a probability of either action or not action. If we augment the original sample $(v^{})$, the model should be able to consistently predict the action region $(l(v^{'}))$. Using spatio-temporal consistency, we propose to bring these predictions close to each other. Firstly, analyzing spatial consistency standalone, we need to evaluate a pixelwise difference between the two predicted localization maps of augmented view $(loc(X^{'}))$ and the original view $(loc(X))$. 

To compare the predictions, we need to inverse the data augmentation for the augmented view $(loc(X^{'}))$ so that mapping between the pixel locations are same while calculating the difference. To minimize this difference in predictions, we use L2 loss. The spatio-temporal consistency loss $(\mathcal{L}_{loc}^{const})$ is defined as,
\begin{equation}
    \mathcal{L}_{loc}^{const} = \mathcal{L}_{L2} = L2(loc(X), (loc(X^{'})^{-1})),
    \label{eqn:locconst}
\end{equation}
where $loc(X^{'})^{-1}$ indicates reversal of augmentations.

The spatio-temporal consistency defined above (\cref{eqn:locconst}) only captures the spatial variance for different predicted localization maps, and, doesn't enforce any temporal constraints. Thus, it effectively works similar to any consistency-based object-detection for images. However, we have a third dimension in videos, the temporal dimension, and moving along this dimension, we can enforce \textit{continuity} and \textit{smoothness} constraints. It means that the predictions should not only be continuous, but the transition across each frame should also be smooth as well.

Therefore, we explore \textit{temporal continuity} of actions in a video to effectively utilize spatio-temporal consistency. We focus on two different aspects of {temporal continuity}, \textit{temporal coherency} and \textit{gradient smoothness}. Temporal coherency captures the relative change in the boundary region of actions across time and helps in refining the detection boundaries. On the other hand, gradient smoothness helps in the detection of abrupt changes in predictions across time.

\paragraph{Temporal coherence}
Temporal coherence is described as the relative displacement of the foreground pixels (action region) in the temporal dimension over a finite amount of frames $(f_{n})$.  We compute the variance of the pixels in the current frame by measuring the relative shift in its position in future and past frames. This pixel-wise variance is computed for all the pixels in a video and is termed as variance map $\mathcal{M}_{var}$. 
The variance map $\mathcal{M}_{var}$ of a video attend to \textit{short-term fine-grained changes} concentrating on the continuity of predictions. Analyzing variance of a particular frame, it will have two distinct regions (\cref{fig:architexture}), \textit{unambiguous}, and \textit{ambiguous}. If a model is confident that a pixel is an action or non-action, we call it \textit{unambiguous} otherwise we describe it as \textit{ambiguous}.  Since the model is already confident on unambiguous regions, we look into the latter. Some of these ambiguous regions will depict the boundaries connecting the foreground and background. Using the variance map we aim to give more \textit{attention} to these regions. This will help the model exploit the ambiguity in spatio-temporal dimensions.

We utilize the variance map as attention to regularize the spatio-temporal consistency loss. This regularized loss $\mathcal{L}_{var}^{const}$ is defined as
\begin{equation}
\footnotesize
    \mathcal{L}_{var}^{const} = w . (\mathcal{M}_{var} \odot \mathcal{L}_{L2}) + (1 - w) . (\mathcal{L}_{L2}),
    \label{eqn:varconst}
\end{equation}
where, mask $\mathcal{M}_{var}$ is calculated as:
\begin{equation}
\footnotesize
    \mathcal{M}_{var} = \frac{\displaystyle\sum_{i=1}^{n}(loc_i - \mu_{n})^2} {n}.
\end{equation}
Here, $loc_{i}$ represents the localization on frame $i$ for which variance is computed, and $n$ represents the total number of frames. $\mu_{n}$ represents the average of $n$ frames. $w$ indicates the weight factor for temporal coherency and non-attentive L2 loss.
However, at the beginning of training, the model will only have primitive knowledge of spatial localization of actions. Therefore, in the initial phase of training, we start with $w=0$ where every pixel in the video has equal importance. As the training progresses, the model can recognize the coarse localization of actions, but, is still unsure of boundary regions. Therefore, we exponentially ramp-up the weight $(w)$ of temporal coherence attention mask $({M}_{var})$ used for L2 loss throughout the training, subsequently, reducing the effect of non-attentive L2 loss. Finally, to exploit longer temporal information, we make use of augmented view. We reversed the spatial augmentation and flip it temporally, attach it to the original view except for the last and first frame and calculate the variance for this longer clip. Since this new clip can be used to make a repetitive cycle, it is termed as \textit{cyclic variance}.

\paragraph{Gradient Smoothness} Taking a deeper look into the temporal aspects of localization, the transition of actor localization should be smooth. To maintain this smoothness constraint, we analyze the change in output localization probability score maps using second-order gradients. Gradient reflects the change in direction. The first-order gradient of a spatio-temporal region along the temporal dimension provides a temporal gradient flow map. Since the offset is small in the temporal dimension, the first-order gradient map should be smooth. Taking the second-order gradient signifies the change in the first-order gradient. As the offset is small, the second-order gradient should be zero. The spikes in the second-order gradient map determine the change in the continuity of the temporal gradient flow map. We utilize this map $\mathcal{M}_{grad}$ as an \textit{attention} to enforce the \textit{long-term smoothness} of spatio-temporal localization. We calculate the gradient smoothness consistency loss as
\begin{equation}
\footnotesize
    \mathcal{L}_{grad}^{const} =(\mathcal{M}_{grad} \odot \mathcal{L}_{L2}),
    \label{eqn:gradconst}
\end{equation}
where mask $\mathcal{M}_{grad}$ is calculated as
\begin{equation}
\footnotesize
    \mathcal{M}_{grad} = \frac{\partial^2 (loc)}{\partial t^2} \text{where} \frac{\partial (loc)}{\partial t} =  \frac{loc_{t+1} - loc_{t-1}}{2}.
    \label{eqn:gradmask}
\end{equation}
Here, the first order partial derivative $\frac{\partial (loc)}{\partial z}$ is approximated using a central difference derivative mask.

\subsection{Overall training objective}
To formalize the final training objective, we have supervised losses and consistency losses. We calculate the supervised loss for classification $(\mathcal{L}_{cls}^{l})$ and localization $(\mathcal{L}_{loc}^{l})$. For consistency, we have classification $(\mathcal{L}_{cls}^{const})$, spatio-temporal $(\mathcal{L}_{loc}^{const})$, temporal coherency $(\mathcal{L}_{var}^{const})$ and gradient smoothness loss $(\mathcal{L}_{grad}^{const})$.
The overall supervised loss is computed as 
\begin{equation}
      \mathcal{L}_{labeled} = \mathcal{L}_{cls}^{l} + \mathcal{L}_{loc}^{l},
      \label{eqn:labeledloss}
\end{equation}
and the combined consistency loss is computed as
\begin{equation}
      \mathcal{L}_{const} = \lambda_{1} \mathcal{L}_{cls}^{const} + \lambda_{2} (\mathcal{L}_{var}^{const}/\mathcal{L}_{grad}^{const}),
      \label{eqn:totalconstloss}
\end{equation}
where $\lambda_{1}$ and $\lambda_{2}$ are weight parameters for classification and spatio-temporal consistency respectively. Finally, the overall training objective is a combination of these two,
\begin{equation}
\label{final_loss}
      \mathcal{L}_{total} = \mathcal{L}_{labeled} +  \mathrm{\lambda}  \mathcal{L}_{const}.
\end{equation}
Here $(\lambda)$ is a weight parameter used for consistency loss.

\section{Experiments}
\label{experiment_analysis}
\paragraph{Datasets} For our action detection experiments, we use UCF101-24 \cite{ucf101} and JHMDB-21 \cite{jhmdb} datasets. \textbf{UCF101-24} contains 3207 untrimmed videos. The number of training and testing videos is 2284 and 923 respectively. It contains 24 action classes. These classes mainly belong to sports and are sub-sampled from the original UCF101 dataset which contains 101 action classes. The original resolution of clips is 320x240. The action duration covers almost 78\% of the total duration of the video. \textbf{JHMDB-21} contains 928 videos categorized into 21 action classes. These classes are similar to sports scenes as UCF. It's a trimmed dataset where the action is happening during the whole video duration. The frame resolution is the same as UCF101-24. To show that our approach can be generalized across other domains, we perform experiments on \textbf{YouTube-VOS} dataset as well. This dataset has 3471 training videos and 589 videos for evaluation. 

\paragraph{Implementation Details} In our experiments, 
we use a frame resolution of 224x224. Our batch size is eight. In each batch, the ratio of labeled to unlabeled samples is 1:1. So, out of eight clips in a batch, four are samples from the labeled subset and the remaining four from the unlabeled subset. Then, they are randomly shuffled. The number of frames per clip is eight. We choose the frames with a skip rate of 2. The distribution of labeled and unlabeled samples for our experiments is 20/80 for UCF101-24 and 30/70 for JHMDB-21 datasets. 
We use I3D \cite{i3d} as a backbone with pretrained weights from Kinetics\cite{k400} and Charades \cite{charades}. 

\paragraph{Training details}
We use an Adam optimizer with an initial learning rate of 1e-4, and a scheduler decay rate of 0.1, if training loss doesn't improve in the last 5 epochs. We train the model for 100 epochs on UCF101-24 and 50 epochs for JHMDB-21. The lambda value for consistency loss is set to 0.1. The parameters $\mathrm{\lambda}_{1}$ and $\mathrm{\lambda}_{2}$ in \cref{eqn:totalconstloss} are set to 0.3 and 0.7. To calculate the temporal coherency of each frame in a clip, 2 future and 2 past frames are picked where we use the localization of the augmented view. This serves as the attention mask to calculate  L2 loss. For gradient smoothness, we calculate the mask values for L2 loss using second order gradient of spatio-temporal prediction on a single clip in the temporal dimension. For labeled samples, we use margin loss \cite{duarte2018videocapsulenet} as classification loss $(\mathcal{L}_{cls}^{l})$ and binary cross-entropy plus dice loss to measure localization loss $(\mathcal{L}_{loc}^{l})$. 

\paragraph{Evaluation metrics}
We compute frame-metric average precision (f-mAP) and video-metric average precision (v-mAP) scores to evaluate the action detection performance. f-mAP calculates the score based on how many frames overlap with the ground truth frames given the IoU, and, v-mAP scores based on video overlapping. We have shown the results for frame-mAP and video-mAP at 0.2 and 0.5.

\paragraph{Baselines}
To compare our work with existing approaches we extend a few semi-supervised image classification approaches to videos. Especially, we looked into pseudo-label \cite{pseudocls}, MixMatch \cite{mixmatch} and Consistency-based Object Detection (Co-SSD (CC)) \cite{co_ssd}. Pseudo-label requires multiple iterations of training, whereas, MixMatch is dependent on stochastic data augmentations. The ratio of labeled to the unlabeled subset is 20 to 80, consistent for all our experiments. 
We use similar augmentation strategy as described in \cite{mixmatch} and generate two views, a weak and a strong one.

\begin{table*}
  \small
  
  \centering
  \resizebox{0.99\linewidth}{!}{
  \begin{tabular}{ cc| cccc |cccc}
    \toprule
    && \multicolumn{4}{c|}{UCF101-24} & \multicolumn{4}{c}{JHMDB-21}\\
    \midrule
     \multicolumn{2}{c|}{Consistency} & \multicolumn{2}{c}{f-mAP} & \multicolumn{2}{c|}{v-mAP} & \multicolumn{2}{c}{f-mAP} & \multicolumn{2}{c}{v-mAP}\\
    \midrule

     CC  & LC & 0.2 & 0.5 & 0.2 & 0.5  & 0.2 & 0.5 & 0.2 & 0.5\\
    \midrule
    
    && 85.1 $\pm$ 0.95 & 61.0 $\pm$ 0.75 & 91.3 $\pm$ 0.25 & 61.1 $\pm$ 1.25 & 87.9 $\pm$ 0.65 & 61.1 $\pm$ 1.40 & 92.5 $\pm$0.85 & 59.5 $\pm$0.20\\
    \midrule
     \checkmark& & 87.8 $\pm$ 0.85 {\textcolor{blue}{($\uparrow$ 2.7)}} & 65.0 $\pm$ 0.90 {\textcolor{blue}{($\uparrow$ 4.0)}} & 93.7 $\pm$ 0.60 {\textcolor{blue}{($\uparrow$ 2.4)}} & 66.2 $\pm$ 1.10 {\textcolor{blue}{($\uparrow$ 5.1)}} &
     
     89.1 $\pm$ 1.32 {\textcolor{blue}{($\uparrow$ 1.2)}} & 62.9 $\pm$ 2.14 {\textcolor{blue}{($\uparrow$ 1.8)}} & 94.1 $\pm$ 0.60 {\textcolor{blue}{($\uparrow$ 1.6)}} & 61.2 $\pm$2.43 {\textcolor{blue}{($\uparrow$ 1.7)}}\\
     
     &  \checkmark&  89.6 $\pm$ 0.30 {\textcolor{green}{($\uparrow$ 4.5)}} & 69.8 $\pm$ 0.05 {\textcolor{green}{($\uparrow$ 8.8)}} & 95.2 $\pm$ 0.15 {\textcolor{green}{($\uparrow$ 3.9)}} & 71.8 $\pm$ 0.05 {\textcolor{green}{($\uparrow$ 10.7)}} & 
     
     89.0 $\pm$ 1.70 {\textcolor{blue}{($\uparrow$ 1.1)}} & 63.4 $\pm$ 1.90 {\textcolor{blue}{($\uparrow$ 2.3)}} & 94.8 $\pm$ 0.60 {\textcolor{green}{($\uparrow$ 2.3)}} & 61.6 $\pm$ 1.70 {\textcolor{blue}{($\uparrow$ 2.1)}} \\
     
      \checkmark& \checkmark  & 89.1 $\pm$ 0.85 {\textcolor{blue}{($\uparrow$ 4.0)}} & 69.5 $\pm$ 0.65 {\textcolor{blue}{($\uparrow$ 8.5)}} & 95.1 $\pm$ 0.30 {\textcolor{blue}{($\uparrow$ 3.8)}} & 71.8 $\pm$ 0.50 {\textcolor{blue}{($\uparrow$ 10.7)}} &
      
      89.2 $\pm$ 2.35 {\textcolor{green}{($\uparrow$ 1.3)}} & 63.6 $\pm$ 2.45 {\textcolor{green}{($\uparrow$ 2.5)}} & 94.4 $\pm$ 0.67 {\textcolor{blue}{($\uparrow$ 1.9)}} & 62.8 $\pm$1.95 {\textcolor{green}{($\uparrow$ 3.3)}} \\

    \bottomrule
    
  \end{tabular}}
  \caption{Performance on UCF101-24 and JHMDB-21 datasets with inclusion of individual and combined consistency losses. The first row indicates supervised training results. Here CC and LC denotes classification and localization consistency.  }
  \label{tab:final_results}
\end{table*}

\subsection{Results}
First, we analyze the classification and spatio-temporal consistency losses. The results are shown in Table \ref{tab:final_results}.

\paragraph{UCF101-24}   From Table \ref{tab:final_results}, we can see, when we apply classification consistency on action features, we get a major improvement in both f-mAP and v-mAP, a boost of 4-5.1\% at 0.5 over the supervised approach. Next, we investigate consistency based on spatio-temporal localization. In general, spatio-temporal consistency outperforms the classification consistency. Especially at v-mAP@0.5, there's a 10.7\% jump in performance over the supervised baseline. This proves that spatio-temporal consistency enforces the network to learn better features. Finally, the combination of both, outperforms the classification by a margin of 4.5-5.6\% at 0.5 metrics, however, relative to spatio-temporal the performance is almost similar. This shows that later has a greater influence than classification consistency.

\paragraph{JHMDB-21} JHMDB-21 is a relatively smaller dataset which leads to a small number of videos per class and the problem of over-fitting. Thus, we examine the performance for different subsets, and, finally, for our work, we used 30\% of the dataset as a labeled subset, which amounts to 189 labeled samples and 471 unlabeled samples. Relative to supervised training on 30\%, we get a boost of approximately 1-2\% for both classification and localization consistency (Table \ref{tab:final_results}). Training with the combination of both consistencies did provide us a gain of roughly 1\% over single consistency on v-mAP@0.5.

We observe in Table \ref{tab:final_results} that combining classification and spatio-temporal consistency doesn't have a significant impact. Since the classification consistency performance is lower as compared to the spatio-temporal, we only rely on spatio-temporal consistency for further experiments.

Next, we analyze the impact of temporal constraints on consistency regularization. From Table \ref{tab:final_results} and \ref{supportfactors}, we evaluate the performance gain for non-attentive L2 versus \textit{temporal coherency} plus non-attentive L2. For UCF101-24, f-mAP and v-mAP at 0.5,  the later outperforms non-attentive L2 by a margin of 0.1 and 0.3\% respectively. For JHMDB-21, we see a better margin of improvement, with a boost of 1\% for f-mAP and 2\% for v-mAP. Coming to \textit{gradient smoothness}, v-mAP at 0.5 for UCF101-24 and JHMDB-21 beat the non-attentive L2 by 0.6\% and 1.5\% respectively. This corroborates our claim that \textit{temporal coherency} and \textit{gradient smoothness} indeed proves out to be effective to enforce temporal continuity constraints.

\begin{table*}
\small
  
  \centering
  \begin{tabular}{c| c c | c c c | c c c}

    \toprule
    \multirow{2}{*}{Method} & \multicolumn{2}{c|}{ \multirow{2}{*}{Backbone}}  & \multicolumn{3}{c|}{UCF101-24}  & \multicolumn{3}{c}{JHMDB-21}  \\
    &&&f-mAP & \multicolumn{2}{c|}{v-mAP} & f-mAP & \multicolumn{2}{c}{v-mAP}\\
    
    \midrule
     & 2-D & 3-D &   0.5  & 0.2 & 0.5 & 0.5 & 0.2 & 0.5 \\
    \midrule
    \textbf{Fully-Supervised} &&&&&&&& \\
    \midrule 
    Singh \etal \cite{Singh_2017_ICCV} $^{\dagger}$ & \checkmark& & -  & 73.5 & 46.3& - & 73.8 & 72.0 \\
    Kalogeitan \etal \cite{kalogeiton_2017} & \checkmark& & 69.5 & 76.5 & 49.2 &  65.7 & 74.2 & 73.7  \\
    Yang \etal \cite{step_cvpr}$^{\dagger}$ & \checkmark& &  75.0 & 76.6 & - &  - & - & - \\
    Song \etal \cite{Song_2019_CVPR}$^{\dagger}$ & \checkmark&& 72.1 & 77.5 & 52.9 &  65.5 & 74.1 & 73.4  \\
    Zhao and Snoek \cite{Zhao_2019_CVPR}$^{\dagger}$ & \checkmark&& - & 78.5 & 50.3& - & - & 74.7 \\
    Li \etal \cite{li2020actions}  & \checkmark&  & 78 & 82.8 & 53.8 & 70.8 & 77.3 & 70.2\\
    Hou \etal \cite{tube_cnn} & & \checkmark& 41.4 & 47.1 & - &61.3 & 78.4 & 76.9  \\
    Gu \etal \cite{ava}$^{\dagger}$ & & \checkmark& 76.3 & - & 59.9& 73.3 & - & 78.6 \\
    Sun \etal \cite{actor_centric} & & \checkmark & - & - & -& \underline{77.9} & - & \underline{80.1} \\
    Pan \etal \cite{pan2021actor} & & \checkmark & \underline{84.3} & -& -& - & - & - \\
    Duarte \etal \cite{duarte2018videocapsulenet} & & \checkmark & 78.6 & \underline{97.1} & \underline{80.3} & 64.6 & 95.1 & -   \\
    Ours && \checkmark& 69.2 & 95.3 & 71.9 & 68.1 & \underline{96.8} & 68.4   \\
    \midrule
    \textbf{Weakly-Supervised} &&&&&&&& \\
    \midrule
    
    Mettes \etal \cite{weakly6} & \checkmark & &- &37.4&-  &- &-&  -\\
    Mettes and Snoek \cite{weakly5} & \checkmark& &- &41.8&- & - &-&  -\\
    Cheron \etal \cite{cheron_2018} & & \checkmark& - & 43.9  & 17.7 &- & - & -  \\
    Escorcia \etal \cite{Escorcia2020GuessWA} & & \checkmark & 45.8 & 19.3 & - & - & - & - \\
    Arnab \etal \cite{Arnab2020UncertaintyAwareWS} & & \checkmark& - & 61.7 & 35.0 & - & - & -  \\
    Zhang \etal \cite{weakly4} & & \checkmark & 30.4 & 45.5 & 17.3& 65.9 & 77.3 & 50.8 \\
    
    \midrule
    \textbf{Semi-Supervised} &&&&&&&& \\
    \midrule
    MixMatch \cite{mixmatch} && \checkmark & 20.2 & 60.2 & 13.8& 7.5 & 46.2 &  5.8 \\
    Psuedo-label \cite{pseudocls} && \checkmark & 64.9&93.0&65.6 & 57.4&90.1&57.4\\
    Co-SSD(CC)\cite{co_ssd} & & \checkmark& 65.3 & 93.7 & 67.5 & 60.7& 94.3& 58.5 \\
    \hline
    Ours  & &    \checkmark&\textbf{ 69.9} & \textbf{95.7} & \textbf{72.1} & \textbf{ 64.4} &\textbf{ 95.4} & \textbf{63.5}   \\
    \bottomrule

  \end{tabular}
  \caption{Comparison with existing supervised and weakly supervised works along with the semi-supervised baselines on UCF101- 24 and JHMDB-21 . $\dagger$ denotes approach uses Optical flow. 
  }
  \label{comparison_sota}
\end{table*}

\subsection{Comparison}
We first compare the proposed approach with the semi-supervised baselines followed by existing works on weakly-supervised learning and supervised learning. This is the first work on semi-supervised video action detection to the best of our knowledge, therefore for a fair comparison, we introduce several standardized semi-supervised baselines. This includes two major sub-areas: consistency (MixMatch, Co-SSD(CC)), and, pseudo-label.

\begin{figure}[t]
     \centering
         \centering
         \includegraphics[width=0.7\linewidth]{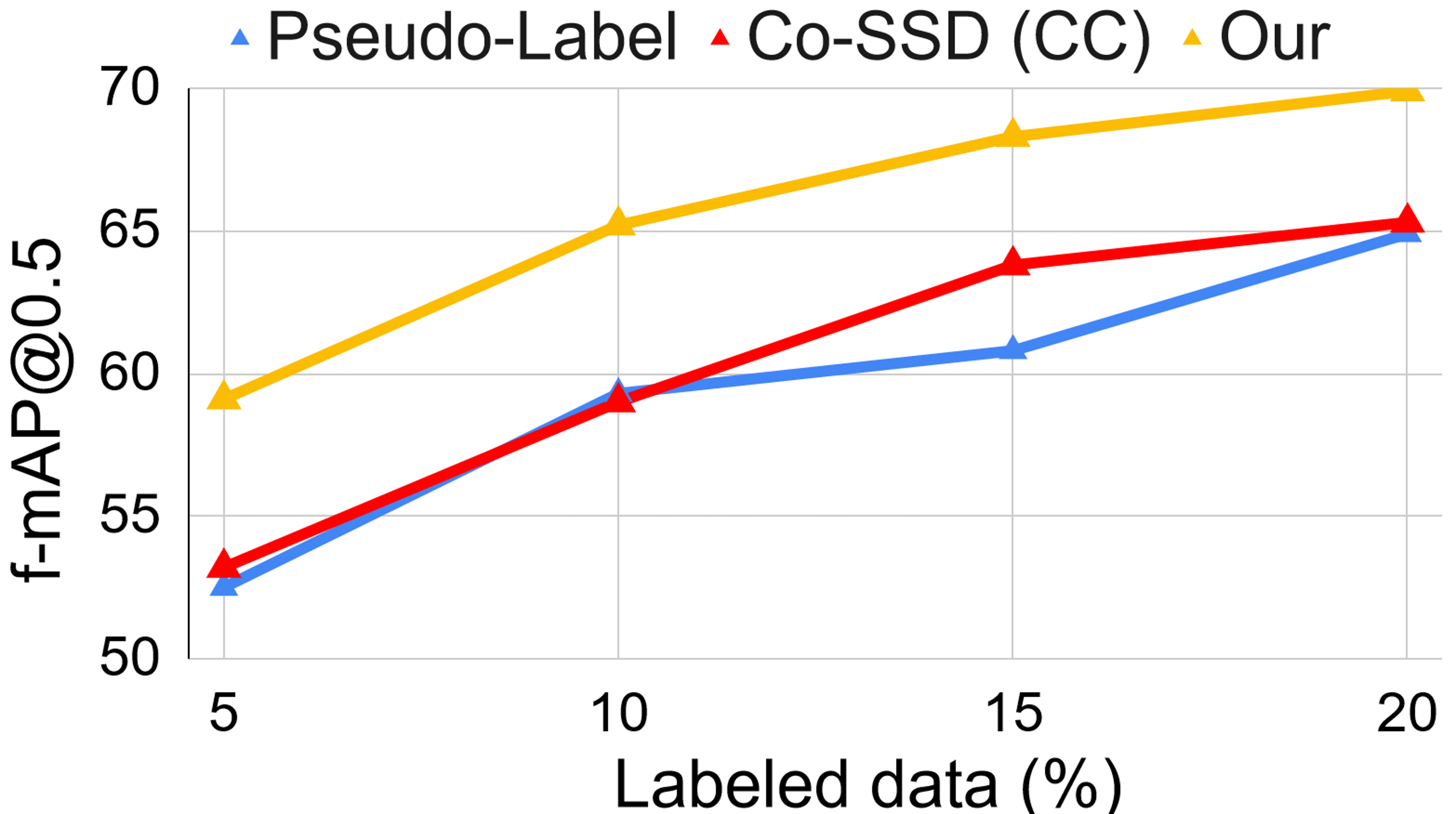}
     \caption{Comparison of pseudo-label \cite{pseudocls}, Co-SSD (CC) \cite{co_ssd}, and ours approach for 5, 10, and, 15 and 20\% of labeled data.}
     \label{fig:comparesemi}
\end{figure}

\paragraph{Semi-supervised} For comparison with semi-supervised approaches, we extended MixMatch, pseudo-label, and Co-SSD to video action detection. The performance of MixMatch is lowest amongst all (Table \ref{comparison_sota}). Compared to pseudo-label at v-mAP@0.5, our approach outperformed by a margin of 5-6\% on UCF101-24 and 7-8\% on JHMDB-21. Co-SSD outperformed the pseudo-label approach, however, we beat that approach with a margin of 4-5\% for both the datasets. We show this comparison in Fig. \ref{fig:comparesemi} for different percentages of labeled samples.

\paragraph{{Weakly-supervised}} These methods \cite{Escorcia2020GuessWA, cheron_2018, Arnab2020UncertaintyAwareWS} use 100\% of the class labels, as well as a state-of-the-art actor detector, to get bounding boxes across the whole video. On the other hand, we did not use any bounding box or class label information for 80\% of the data. We beat the best-reported scores on UCF101-24 by a margin of more than 35\% approximately. A comparison is shown in Table \ref{comparison_sota}.

\paragraph{{Supervised}} Table \ref{comparison_sota} shows a comparison with several existing supervised action detection approaches. We observe that with only 20\% labeled data for UCF101-24, our scores at v-mAP@0.2 and 0.5 outperform all of the approaches. f-mAP@0.5 is better than most of the approaches apart from  \cite{step_cvpr}, \cite{Song_2019_CVPR}, and, \cite{ava}. However, all of them have used optical flow as a second modality. Optical flow works as an extra supervisory signal. For JHMDB-21, we were able to beat some of the approaches at f-mAP@0.5 and v-mAP@0.2.

\section{Ablation Study}
To get a deeper insights on how \textit{attention} constraint helped in improving the accuracy, we did a study on individual components of \textit{temporal coherence} and \textit{gradient smoothness} modules. Since JHMDB-21 is a small dataset, we analyze the results for three different seed variations for each of the components on JHMDB-21. The ablation scores are shown in Table \ref{supportfactors}. For JHMDB, the scores are mean of three runs. 

\begin{table}

  \centering
  \resizebox{0.99\linewidth}{!}{
  \begin{tabular}{cccc | cc | cc}
    \toprule
     \multicolumn{4}{c|}{Experiment}  & 
     \multicolumn{2}{c|}{UCF101-24} & \multicolumn{2}{c}{JHMDB-21}\\ 
    \midrule
    V & G & VC & L2    & f-mAP@0.5  & v-mAP@0.5 & f-mAP@0.5  & v-mAP@0.5    \\
    \midrule
    \midrule
     \checkmark &&& & 68.3   & 70.3 & 61.9 & 61.4 \\

    & & \checkmark & &  68.4 \scriptsize{\textcolor{blue}{($\uparrow$ 0.1)}}  & 70.8 \scriptsize{\textcolor{blue}{($\uparrow$ 0.5)}} & 63.0 \scriptsize{\textcolor{blue}{($\uparrow$ 1.1)}} & 61.5 \scriptsize{\textcolor{blue}{($\uparrow$ 0.1)}} \\

    \checkmark && &\checkmark & 68.8 \scriptsize{\textcolor{blue}{($\uparrow$ 0.5)}} &  71.6 \scriptsize{\textcolor{blue}{($\uparrow$ 1.3)}} & 63.3 \scriptsize{\textcolor{blue}{($\uparrow$ 1.4)}}& 62.4 \scriptsize{\textcolor{blue}{($\uparrow$ 1.0)}} \\
    
    & &\checkmark& \checkmark  & \textbf{69.9} \scriptsize{\textcolor{blue}{($\uparrow$ 1.6)}}  &  \textbf{72.1} \scriptsize{\textcolor{blue}{($\uparrow$ 1.8)}} &  \textbf{64.4} \scriptsize{\textcolor{blue}{($\uparrow$ 2.5)}} &\textbf{ 63.5} \scriptsize{\textcolor{blue}{($\uparrow$ 2.1)}}\\
    \midrule
    &\checkmark & & & 69.6  & 72.4 & 63.2 & 63.1\\
    
    &\checkmark & & \checkmark & 69.4 \scriptsize{\textcolor{red}{($\downarrow$ 0.2)}}  & 72.0 \scriptsize{\textcolor{red}{($\downarrow$ 0.4)}} & 63.1 \scriptsize{\textcolor{red}{($\downarrow$ 0.1)}} & 62.2 \scriptsize{\textcolor{red}{($\downarrow$ 0.9)}}\\
    
    \bottomrule
   
  \end{tabular}}
   \caption{An analysis of the effect of temporal constraints on consistency regularization using UCF101 - 20\%  and JHMDB-21 - 30\% labeled subset. 
   V, G, VC and L2 stands for Variance, Gradient, Cyclic variance and non-attentive L2 loss.
   }
   \label{supportfactors}
\end{table}

\textbf{\textit{Temporal coherence:}}  From Table \ref{supportfactors}, when we apply only attention mask, we outperform the supervised baseline by a good margin. Using cyclic variance improved the score by 0.1-0.5\% for UCF101-24 and 0.1-1.1\% for JHMDB-21 dataset. Incorporating non-attentive L2 loss, we see a good increment of roughly 1\% for both datasets. Finally, moving onto the cyclic variance alongwith non-attentive L2 we got an additional boost of 1\%. This demonstrates that not only temporal variance is helping, but longer temporal information (cyclic variance) also compliments the base score as well. We also notice that the margin of improvement is higher when we are utilizing a pixel-wise prediction mask (JHMDB-21).

\textbf{\textit{Gradient smoothness:}} Following the paths of temporal coherence, we first employed gradient smoothness with non-attentive L2. However, this did not improve the score further as compared to standalone gradient smoothness loss. Using only that, we see it outperforms non-attentive L2 on v-mAP@0.5 for both datasets. This was expected as \textit{gradient smoothness} focuses on the whole clip compared to \textit{temporal coherency}.

\section{Discussions}
In this section, we discuss some of the queries pertaining to semi-supervised activity detection in general. 

\begin{figure}[t]
     \centering
         \centering
         \includegraphics[width=0.48\linewidth]{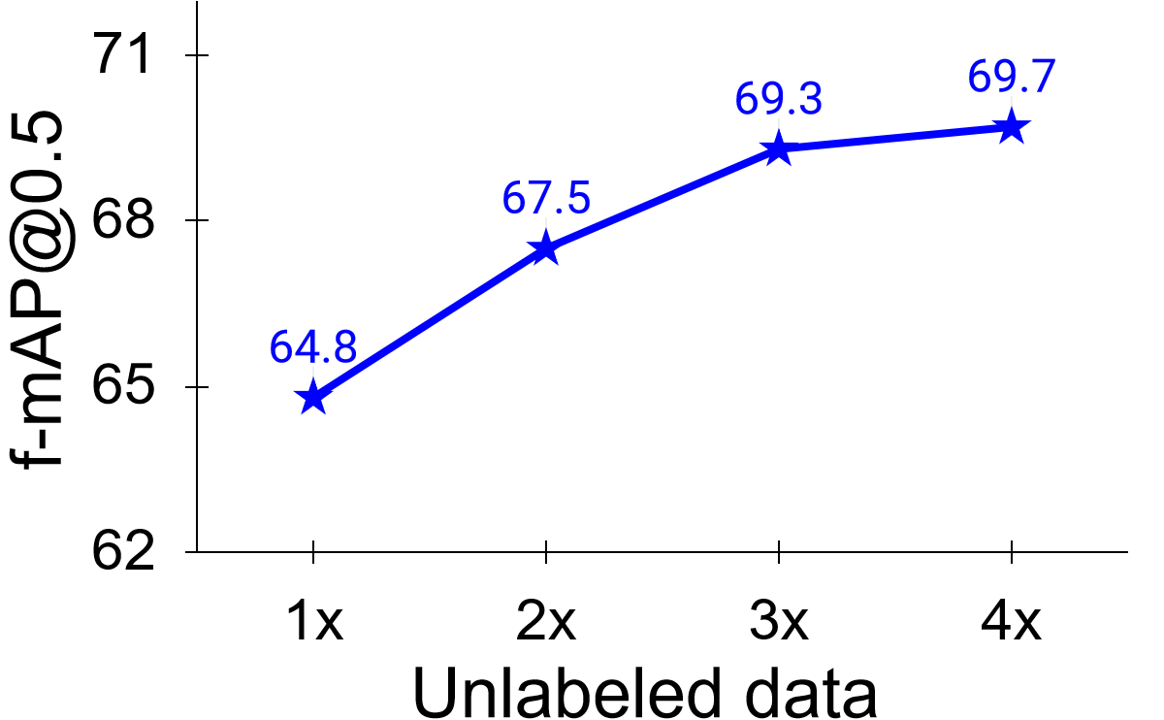}
         \includegraphics[width=0.48\linewidth]{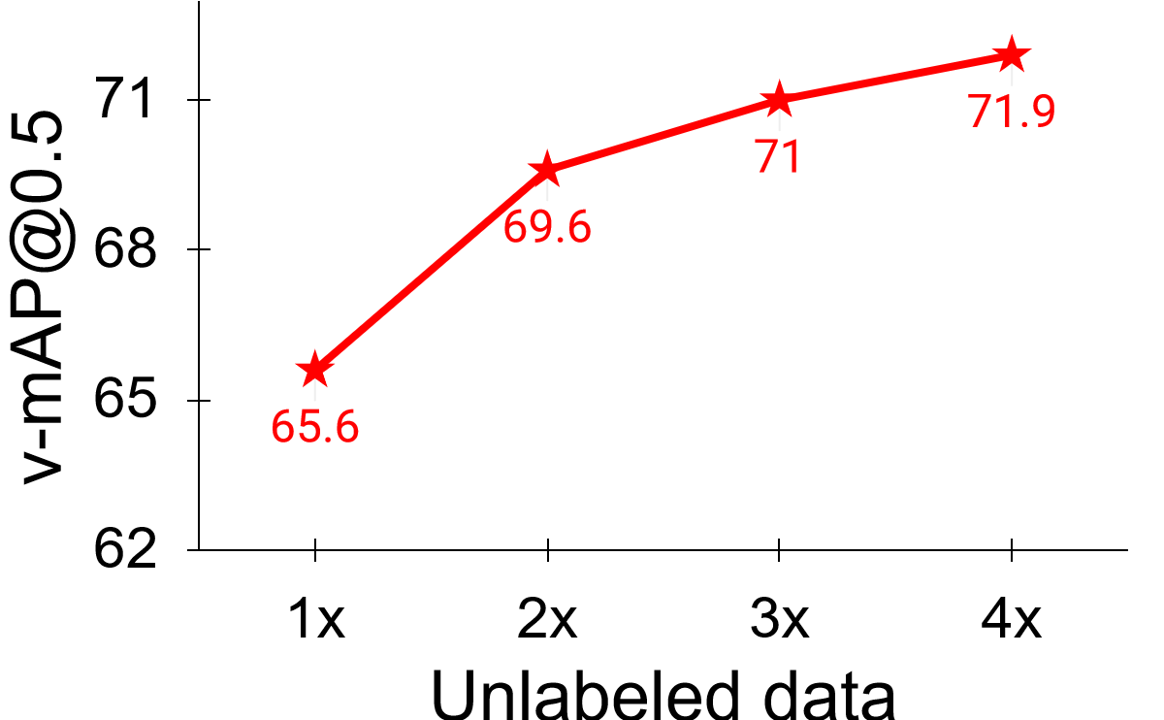}
     \caption{Performance with varying the amount of unlabeled data.}
  \label{fig:unlabeldata_amplify}
\end{figure}

\textit{\textbf{Does the amount of unlabeled samples matters?}}
For this study, we keep the amount of labeled samples constant to 20\%. Then, we increase the amount of unlabeled samples from 20\% (1x) to 80\% (4x). (Fig. \ref{fig:unlabeldata_amplify}) We see a constant gain in performance for all the metrics. This shows that more the number of unlabeled samples, better will be the performance.

\begin{table}
  \small
  
  \centering
  
  \begin{tabular}{ccccc}
    \toprule
    \multirow{2}{*}{Dataset} & \multicolumn{2}{c}{f-mAP (\%)} & \multicolumn{2}{c}{v-mAP (\%)}\\
        & 0.2 & 0.5 & 0.2 & 0.5  \\ \midrule 
        \midrule

     Trimmed &  90.2 & 69.9 & 96.2 & 72.1 \\
     Untrimmed  & 90.1& 69.5 & 96.0  &71.3 \\
     \bottomrule
  \end{tabular}
  \caption{Performance comparison on UCF101-24 dataset using untrimmed videos instead of trimmed where the action is occurring in all the video frames. We observe a small to negligible performance drop which indicates that our method can also utilize untrimmed videos.}
  \label{data_random}
  
\end{table}

\begin{figure}[t]
     \centering
         \centering
         \includegraphics[width=0.7\linewidth]{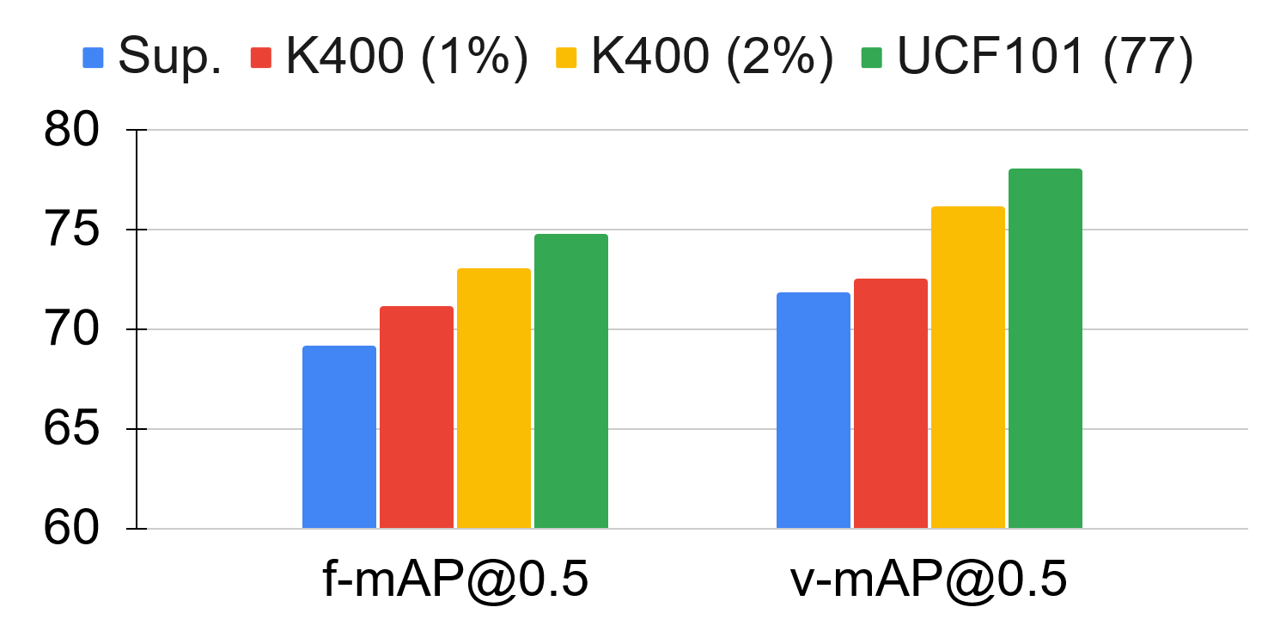}
     \caption{We observe performance gain when additional unlabeled videos were used from external sources, such as Kinetic-400 (K400) and UCF-101 (other 77 action classes from UCF-101).}
  \label{fig:extra_data}
\end{figure}

\textbf{\textit{What is the impact of using an untrimmed dataset instead of trimmed?}} In trimmed videos, all the frames in a video sample contain action, however, in untrimmed, there may be some frames without any actions. In our experiments, we assume the availability of trimmed videos for UCF101-24. We conduct experiments with a scenario where the unlabeled videos can be untrimmed. The dataset has some videos with no activity at all. The evaluation is shown in Table \ref{data_random}. 
We only observe a marginal to negligible drop in performance
which shows the robustness of the proposed spatio-temporal consistency for untrimmed videos. This is also evident from our next set of experiments where we utilize additional untrimmed unlabeled videos from external datasets.

\textit{\textbf{Can additional data support as a supervisory signal?}} Lastly, we explore how can we utilize the rest of the actions (not labeled) from UCF101 and outperform the supervised accuracy. In Fig. \ref{fig:extra_data} we observe that additional video samples even from the Kinetics dataset help in improving the performance. However, the gain is more significant when the videos are from a similar distribution, UCF-101 in this scenario.

\begin{table}
  
  \small
  \centering
  \resizebox{0.9\linewidth}{!}{
  \begin{tabular}{ccccccc}
   \toprule
    Method & Semi-Sup. & Avg & $J_{S}$ & $J_{U}$ & $F_{S}$ & $F_{U}$ \\
    \midrule
    LSTM\cite{voslstm}  &   & 10.1 & 11.6 & 10.1 & 9.6 & 9.2 \\
     & \checkmark & 36.8 & 43.1 & 31.4 & 40.8 & 31.8\\
    \midrule
     Sup. (100\%)& & 47.9 & 55.7 & 39.6 & 55.2 &41.3 \\
    \bottomrule
  \end{tabular}}
  \caption{Evaluation on Youtube-VOS dataset. We have used 10\% data for supervised approach. The bottom row shows the results for supervised training on 100\% data. }
  \label{vos_experiments}
\end{table}

\subsection{Generalization to video object segmentation}

We have also shown that our approach is generalizable across different task. For VOS dataset \cite{vosdataset}, we see an overall improvement of 30\% over the supervised baseline  (Table \ref{vos_experiments}). We have shown the score using \textit{temporal coherency} consistency loss. 

\section{Conclusion}

In this work, we propose a novel end-to-end approach for \textit{semi-supervised video action detection}. To best of our knowledge, this is the \textit{first attempt} in semi-supervised learning for action detection. We propose the use of \textit{consistency regularization} for an efficient and effective detection performance. We demonstrate the positive impact of \textit{temporal coherency} and \textit{gradient smoothness} constraint for spatio-temporal localization. The proposed approach achieves significant performance boost over \textit{supervised baselines} with limited labels and \textit{outperforms} weakly supervised methods.

{\small
\bibliographystyle{ieee_fullname}
\bibliography{main}
}

\end{document}